\documentclass[10pt,twocolumn,letterpaper]{article}

\usepackage{cvpr}
\usepackage{times}
\usepackage{epsfig}
\usepackage{graphicx}
\usepackage{amsmath}
\usepackage{amssymb}

\usepackage{amsmath}
\usepackage{graphics}
\usepackage{graphicx}
\usepackage{amssymb}
\usepackage{algorithm}
\usepackage{algorithmic}
\usepackage{mathrsfs}
\usepackage{booktabs}
\usepackage{bm}
\usepackage{bbm}
\usepackage{float}
\usepackage{url}
\usepackage{amsfonts}
\usepackage[square,sort&compress,numbers]{natbib}
\usepackage{multirow}
\usepackage{textcomp}
\usepackage{supertabular}
\usepackage{array}
\usepackage{caption}
\usepackage{gensymb}

\usepackage[breaklinks=true,bookmarks=false]{hyperref}

\cvprfinalcopy 

\def\cvprPaperID{5946} 

\ifcvprfinal\fi
\begin{document}

\urlstyle{rm}

\newcolumntype{L}[1]{>{\raggedright\arraybackslash}p{#1}}
\newcolumntype{C}[1]{>{\centering\arraybackslash}p{#1}}
\newcolumntype{R}[1]{>{\raggedleft\arraybackslash}p{#1}}

\title{Stylized Neural Painting}

\author{
Zhengxia Zou$^1$, \ \ Tianyang Shi$^2$,  \ \ Shuang Qiu$^1$, \ \ Yi Yuan$^2$,  \ \ Zhenwei Shi$^3$\\
$^1$University of Michigan, Ann Arbor, \ \ $^2$NetEase Fuxi AI Lab, \ \ $^3$Beihang University \\
}

\twocolumn[{%
\renewcommand\twocolumn[1][]{#1}%
\maketitle
\centering{\includegraphics[width=1.0\linewidth]{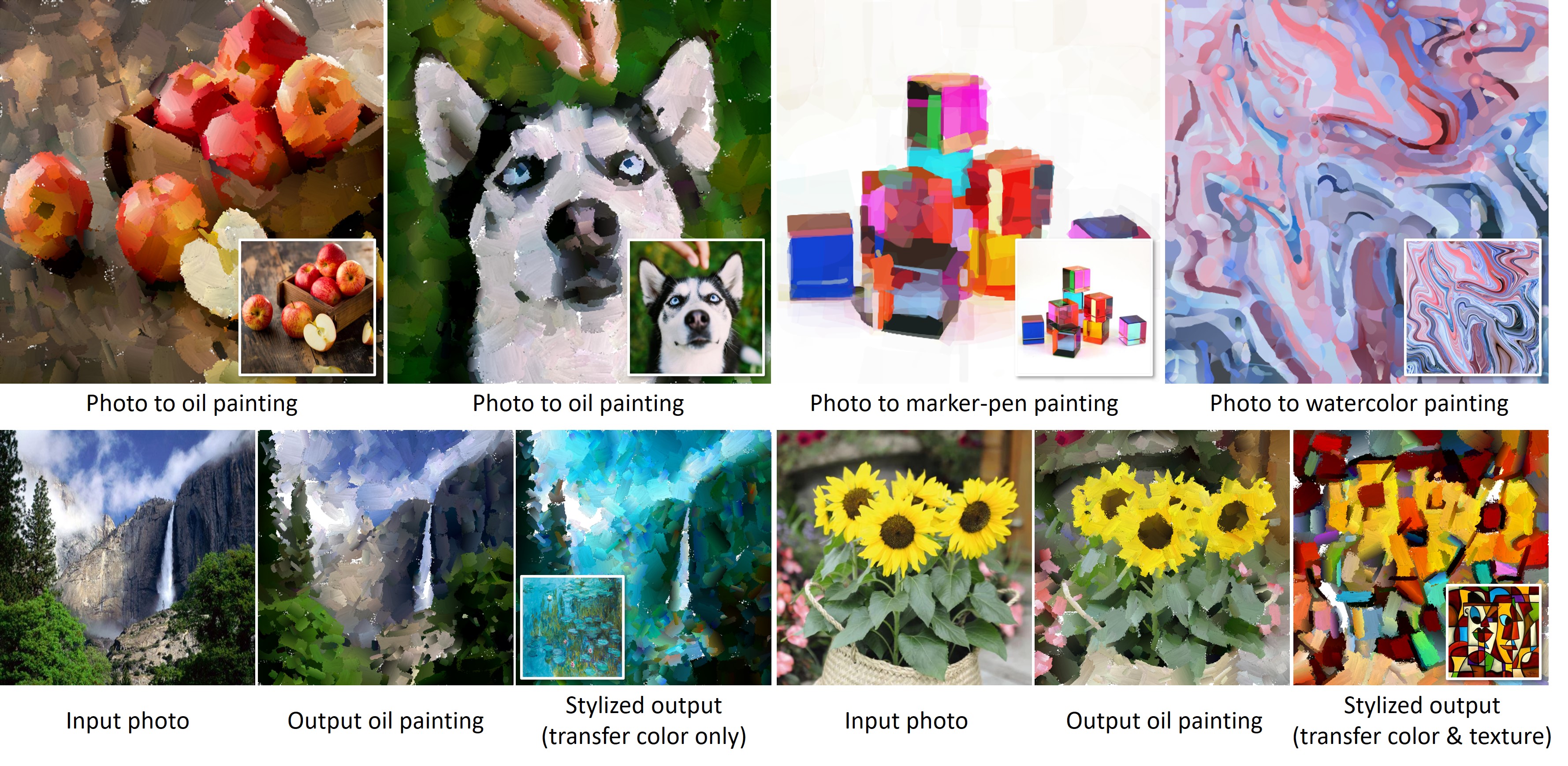}}
\captionof{figure}{\small We propose a stroke-rendering based method that generates realistic painting artworks. The paintings are generated with a vector format, which can be further optimized and rendered with different visual styles. We recommend zooming in to see the stroke textures.}
\label{fig:teaser}
\vspace{2em}
}]

\begin{abstract}
\vspace{-0.5em}
This paper proposes an image-to-painting translation method that generates vivid and realistic painting artworks with controllable styles. Different from previous image-to-image translation methods that formulate the translation as pixel-wise prediction, we deal with such an artistic creation process in a vectorized environment and produce a sequence of physically meaningful stroke parameters that can be further used for rendering. Since a typical vector render is not differentiable, we design a novel neural renderer which imitates the behavior of the vector renderer and then frame the stroke prediction as a parameter searching process that maximizes the similarity between the input and the rendering output. We explored the zero-gradient problem on parameter searching and propose to solve this problem from an optimal transportation perspective. We also show that previous neural renderers have a parameter coupling problem and we re-design the rendering network with a rasterization network and a shading network that better handles the disentanglement of shape and color. Experiments show that the paintings generated by our method have a high degree of fidelity in both global appearance and local textures. Our method can be also jointly optimized with neural style transfer that further transfers visual style from other images. Our code and animated results are available at \url{https://jiupinjia.github.io/neuralpainter/}.

\end{abstract}

\vspace{-1em}
\section{Introduction}

Creating artistic paintings is one of the defining characteristics of humans and other intelligent species. In recent years, we saw great advancements in generative modeling of image translation or style transfer which utilizes neural network as a generative tool~\cite{zhu2017unpaired,gatys2016image,mordvintsev2015inceptionism,johnson2016perceptual}. Previous image-to-image translation and style transfer methods typically formulate the translation either as a pixel-wise mapping~\cite{zhu2017unpaired,johnson2016perceptual} or a continuous optimization process in their pixel space~\cite{gatys2016image}. However, as an artistic creation process, the paintings usually proceed as a sequentially instantiated process that creates using brushes, from abstract to concrete, and from macro to detail. This process is fundamentally different from how neural networks create artwork that produces pixel-by-pixel results. To fully master the professional painting skills, people usually need a lot of practice and learn domain expertise. Even for a skilled painter with years of practice, it could still take hours or days to create a realistic painting artwork.

In this paper, we explore the secret nature of human painting and propose an automatic image-to-painting translation method that generates vivid and realistic paintings with controllable styles. We refer to our method as ``Stylized Neural Painter''. Instead of manipulating each of the pixels in the output image, we simulate human painting behavior and generate vectorized strokes sequentially with a clear physical significance. Those generated stroke vectors can be further used for rendering with arbitrary output resolution. Our method can ``draw'' in a variety of painting styles, e.g. oil-painting brush, watercolor ink, marker-pen, and tape art. Besides, our method can also be naturally embedded in a neural style transfer framework and can be jointly optimized to transfer its visual style based on different style reference images.

In our method, different from the previous stroke-based rendering methods that utilize step-wise greed search~\cite{haeberli1990paint,litwinowicz1997processing}, recurrent neural network~\cite{ha2017neural}, or reinforcement learning~\cite{xie2013artist,ganin2018synthesizing,huang2019learning,zhou2018learning}, we reformulate the stroke prediction as a ``parameter searching'' process that aims to maximize the similarity between the input and the rendering output in a self-supervised manner. Considering that a typical graphic render is not differentiable, we take advantage of the neural rendering that imitates the behavior of the graphic rendering and make all components in our method differentiable. We show that previous neural stroke renderers~\cite{huang2019learning,shi2019face,shi2020neural} may suffer from the parameter coupling problem when facing complex rendering scenarios, e.g., brushes with real-world textures and color-transition. We, therefore, re-design the neural renderer and decomposed the rendering architecture into a rasterization network and a shading network, which can be jointly trained and rendered with much better shape and color fidelity. We also found interestingly that the pixel-wise similarity like $\ell_1$ or $\ell_2$ pixel loss, may have an intrinsic flaw of zero-gradient on optimizing over the vectorized parameters, although these losses have been widely used in a variety of image translation tasks~\cite{zhu2017unpaired,johnson2016perceptual,ledig2017photo}. We show that this problem lies in the different nature of stroke parameterization and rasterization, and propose to solve this problem from the perspective of optimal transportation. Specifically, we consider the movement of a stroke from one location to another as a transportation process, where we aim to minimize the efforts of that movement.

We test our method on various real-world images and photos, including human portraits, animals, scenery, daily objects, art photography, and cartoon images. We show that our method can generate vivid paintings with a high degree of realism and artistic sense in terms of both global visual appearance and local texture fidelity.

The contribution of our paper is summarized as follows:
\begin{itemize}
\item We propose a new method for stroke based image-to-painting translation. We re-frame the stroke prediction as a parameter searching processing. Our method can be jointly optimized with neural style transfer in the same framework.
\vspace{-0.5em}
\item We explore the zero-gradient problem on parameter searching and view the stroke optimization from an optimal transport perspective. We introduce a differentiable transportation loss and improves stroke convergence as well as the painting results.
\vspace{-0.5em}
\item We design a new neural renderer architecture with a dual-pathway rendering pipeline (rasterization + shading). The proposed renderer better deals with the disentanglement of the shape and color and outperforms previous neural renderers with a large margin.
\end{itemize}

\section{Related Work}

{\bf Image translation and style transfer.} Image translation, which aims at translating images from one domain (e.g., real photos) to another (e.g., artworks), has drawn great attention in recent years. GAN based image translation such as Pix2Pix~\cite{isola2017image}, CycleGANs~\cite{zhu2017unpaired} and their variants~\cite{park2019semantic} has played an important role in tasks like image synthesis, semantic editing, style transfer, and also has been applied to computer-generated arts~\cite{abbott2019creating}. In addition to the GAN based method, neural style transfer has also made breakthroughs in stylized image synthesis and is widely used for artwork creation~\cite{johnson2016perceptual, gatys2016image}. Besides, the Deep Dream~\cite{mordvintsev2015inceptionism}, a method that was initially designed to help visualize deep neural networks, also has become a new form of psychedelic and abstract art. Despite the above applications, these methods all generate paintings in a pixel-by-pixel manner, which deviates from the fact that humans use brushes to paint.

{\bf Differentiable rendering.} Rendering is a fundamental problem in computer graphics that converts 3D models into 2D images. Traditional rendering pipelines typically involve a discrete operation called rasterization, which makes the rendering non-differentiable. Differentiable rendering~\cite{loper2014opendr, kato2018neural,Liu_2019_ICCV, li2018differentiable} breaks such limitations and allows calculation of the derivative from the rendering output to the input parameters such as shape, camera pose, and lighting. Since deep neural networks are naturally differentiable in their topology, a new research topic called ``neural rendering'' quickly emerged~\cite{eslami2018neural,nguyen2018rendernet,shi2019face}, which bridges the gap between graphic rendering and deep neural networks.

\textbf{Image sketching/painting.} Humans are born with the ability to abstract and sketch object instances. Early methods on image sketching deal with this problem by using step-wise greedy search or require user interaction~\cite{hertzmann2003survey}. Recent approaches typically train recurrent neural networks~\cite{ha2017neural} and Reinforcement Learning (RL) agent~\cite{xie2013artist,zhou2018learning,ganin2018synthesizing}, or integrate adversarial training~\cite{nakano2019neural} to generate non-deterministic stroke sequences. More recently, thanks to the recent advances of neural rendering~\cite{eslami2018neural,nguyen2018rendernet}, computers are now able to generate more realistic painting artworks~\cite{nakano2019neural,huang2019learning,ganin2018synthesizing}. Among these methods, ``Learning to Paint''~\cite{zhou2018learning} has a similar research motivation to ours, both dedicated to generating stroke based realistic paintings. However, our method differs from this method in several aspects. First, this method generates strokes by using RL while we formulate this process as stroke parameter searching since training LR agents is computationally expensive. Second, we focus on stroke-based style transfer, which is a rare studied problem. Finally, also we redesign the neural renderer and introduce optimal transport methods to this problem.

\begin{figure}
    \centering{\includegraphics[width=1.0\linewidth]{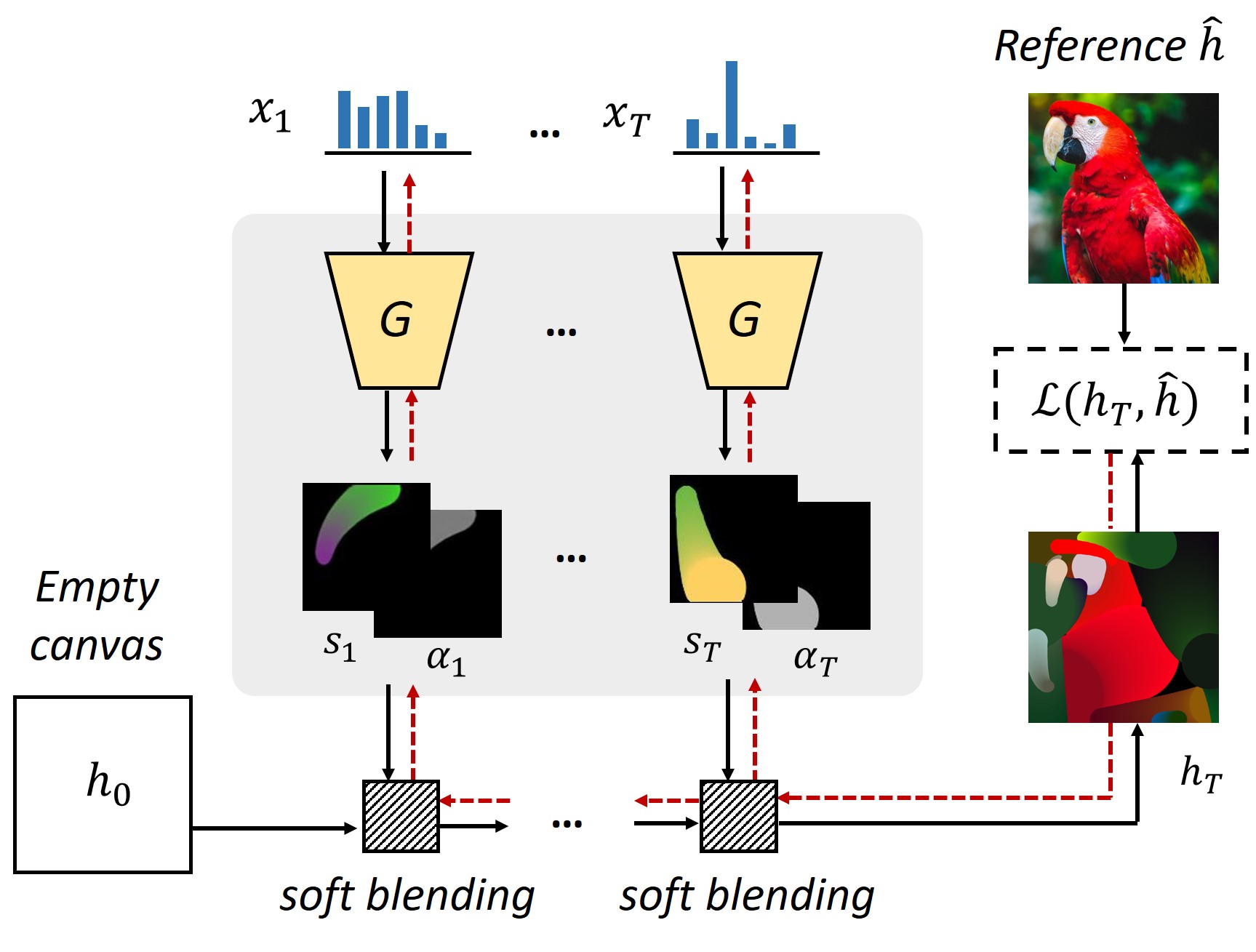}}\\
    \caption{We start from an empty canvas and then render stroke-by-stroke with soft blending. We use gradient descent to find a set of ``optimal'' stroke parameters that minimize the loss $\mathcal{L}$. Here black arrowlines mean forward propagation and red ones mean back-propagation of the gradient.}
    \label{fig:rendering_pipeline}
\end{figure}

\section{Methodology}

Our method consists of three technical modules: 1) a neural renderer that is trained to generate strokes given a set of vectorized stroke parameters; 2) a stroke blender that combines multiple rendering strokes in a differentiable manner; 3) a similarity measurement module that enforces the reconstruction of the input image. In the following, we introduce how each module works accordingly.

\subsection{Overview}

Fig.~\ref{fig:rendering_pipeline} shows an overview of our method. Given an empty canvas $h_0$, we draw step-by-step and superimpose those strokes rendered at each step iteratively. In each drawing step $t$, a trained neural renderer $G$ takes in a set of stroke parameters $\bm{x}_t$ (e.g., shape, color, transparency, and texture), and produces a stroke foreground $s_t$ and an alpha matte $\alpha_t$. We then use soft blending to mix the canvas, the foreground, and alpha matte at each step $t$ and make sure the entire rendering pipeline is differentiable. The soft blending is defined as follows:
\begin{equation}
    h_{t+1} = \alpha_t s_t + (1-\alpha_t)h_t,
\end{equation}
where $(s_t, \alpha_t) = G(\bm{x}_t)$. We finally gather the stroke parameters from all the $T$ steps and optimize them by searching within the stroke parameter space. The searching is conducted under a self-supervised manner, i.e., we enforce the final rendered output $h_T$ similar to a reference image $\hat{h}$:
\begin{equation}
    h_{T} = f_{t=1\sim T}(\Tilde{\bm{x}}) \approx \hat{h},
\end{equation}
where $f_{t=1\sim T}(\cdot)$ is a recursive mapping from stroke parameters to the rendered canvas. $\Tilde{\bm{x}}=[\bm{x_1}, ..., \bm{x}_T]$ are the collection of stroke parameters at $t=1,2,...T$ drawing steps.

Suppose $\mathcal{L}$ represents a loss function that measures the similarity between the canvas $h_T$ and the reference $\hat{h}$, we optimize all the input strokes $\Tilde{\bm{x}}$ at their parameter space and minimize the facial similarity loss $\mathcal{L}$: $\Tilde{\bm{x}^\star} = \arg \min_{\Tilde{\bm{x}}} \mathcal{L}(h_T, \hat{h})$. We use gradient decent to update the strokes as follows:
\begin{equation}
    \Tilde{\bm{x}} \leftarrow \Tilde{\bm{x}} - \mu \frac{\partial \mathcal{L}(h_T, \hat{h})}{\partial \Tilde{\bm{x}}},
\end{equation}
where $\mu$ is a predefined learning rate.

\begin{figure}
    \centering{\includegraphics[width=1.0\linewidth]{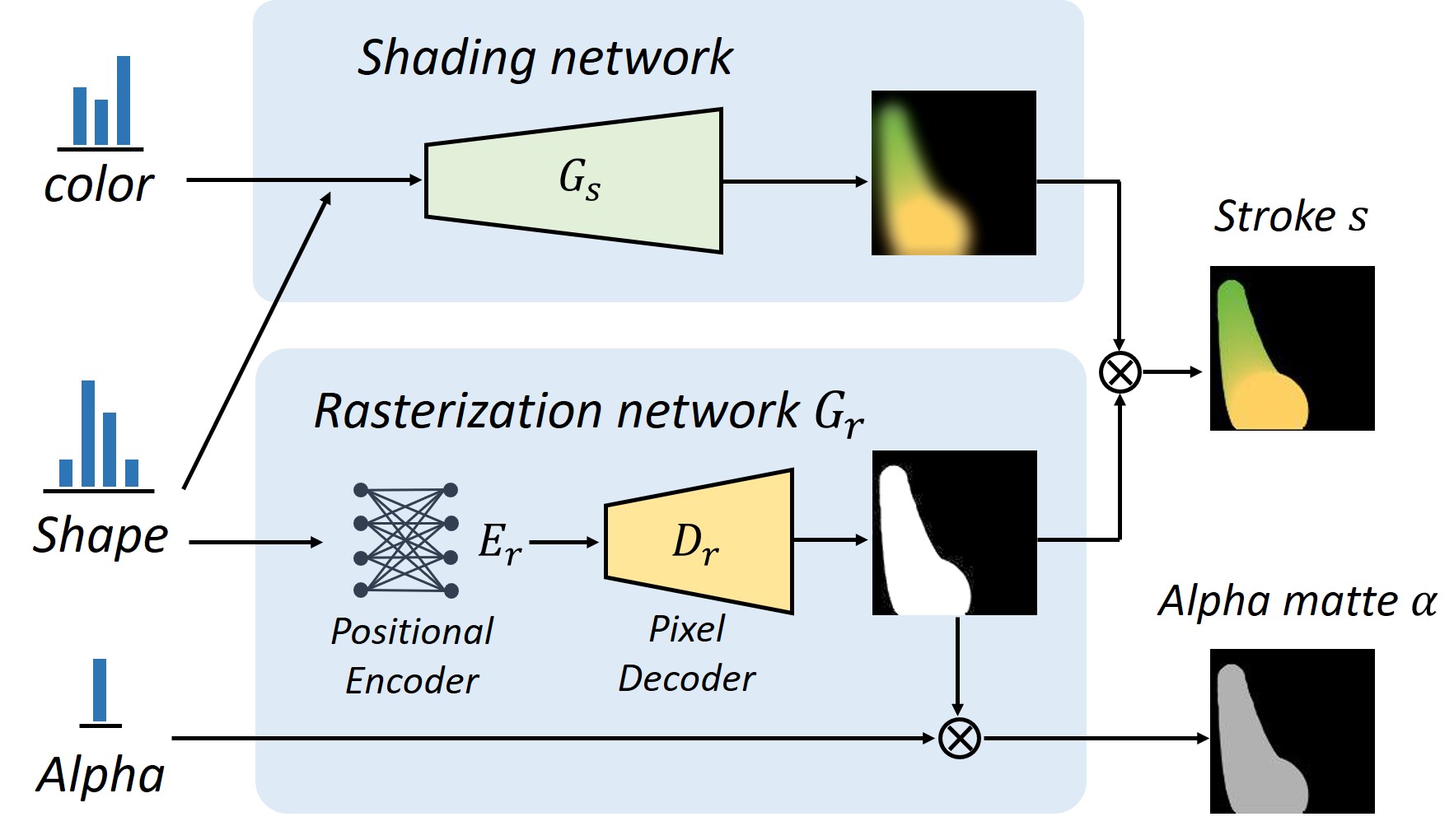}}\\
    \caption{We design a dual-pathway neural renderer which consists of a shading network $G_s$ and a rasterization network $G_r$. Our renderer takes in a group of stroke parameters (color, shape, and transparency) and produces the rasterized foreground map and alpha matte.}
    \label{fig:renderer}
\end{figure}

\subsection{Disentangle neural rendering}

To build a neural renderer, a general practice is to build a deep convolutional network and train it to imitate the behavior of a graphic engine. Previous researches proposed to use stacked transposed convolutions~\cite{shi2019face,zhu2016generative} or position encoder +  decoder architecture~\cite{zheng2018strokenet,huang2019learning}. These approaches can work well in simple stroke rendering scenarios. However, we find these renderers may suffer from a coupling of shape and color representations when testing with more complex rendering settings like color transition and stroke textures. We propose to solve this problem by designing a dual-pathway neural renderer that disentangles color and shape/texture through the rendering pipeline.

As shown in Fig.~\ref{fig:renderer}, the proposed neural renderer consists of two networks, a shading network $G_s$ and a rasterization network $G_r$. We divide the parameters of a stroke $\bm{x}$ into three groups: color, shape, and transparency. We build $G_s$ as a stack of several transposed convolution layers~\cite{zeiler2014visualizing}, which takes in both the color and shape parameters and generates strokes with faithful foreground color. We design the $G_r$ as a positional encoder + a pixel decoder, which simply ignores the color but generates stroke silhouette with a clear shape boundary. We finally generate the output stroke foreground $s$ by masking the color map with the stroke silhouette and generate the final alpha matte $\alpha$ by rescaling the silhouette using the input alpha value.

We train our neural renderer with standard $\ell_2$ pixel regression losses on both the rendered stroke foreground and the alpha matte. During the training, we minimize the following objective function:
\begin{equation}\label{eq:G_loss}
    \mathcal{L}_G (\bm{x}) = \mathbb{E}_{\bm{x}\sim u(\bm{x})}\{ {\|s - \hat{s}\|}_2^2 + {\|\alpha - \hat{\alpha}\|}_2^2 \},
\end{equation}
where $\hat{s}$ and $\hat{\alpha}$ are the ground truth foreground and alpha matte rendered by the graphic engine. $\bm{x}\sim u(\bm{x})$ are stroke parameters randomly sampled within their parameter space.

\subsection{Pixel similarity and zero-gradient problem}

\begin{figure}[t]
    \centering{\includegraphics[width=\linewidth]{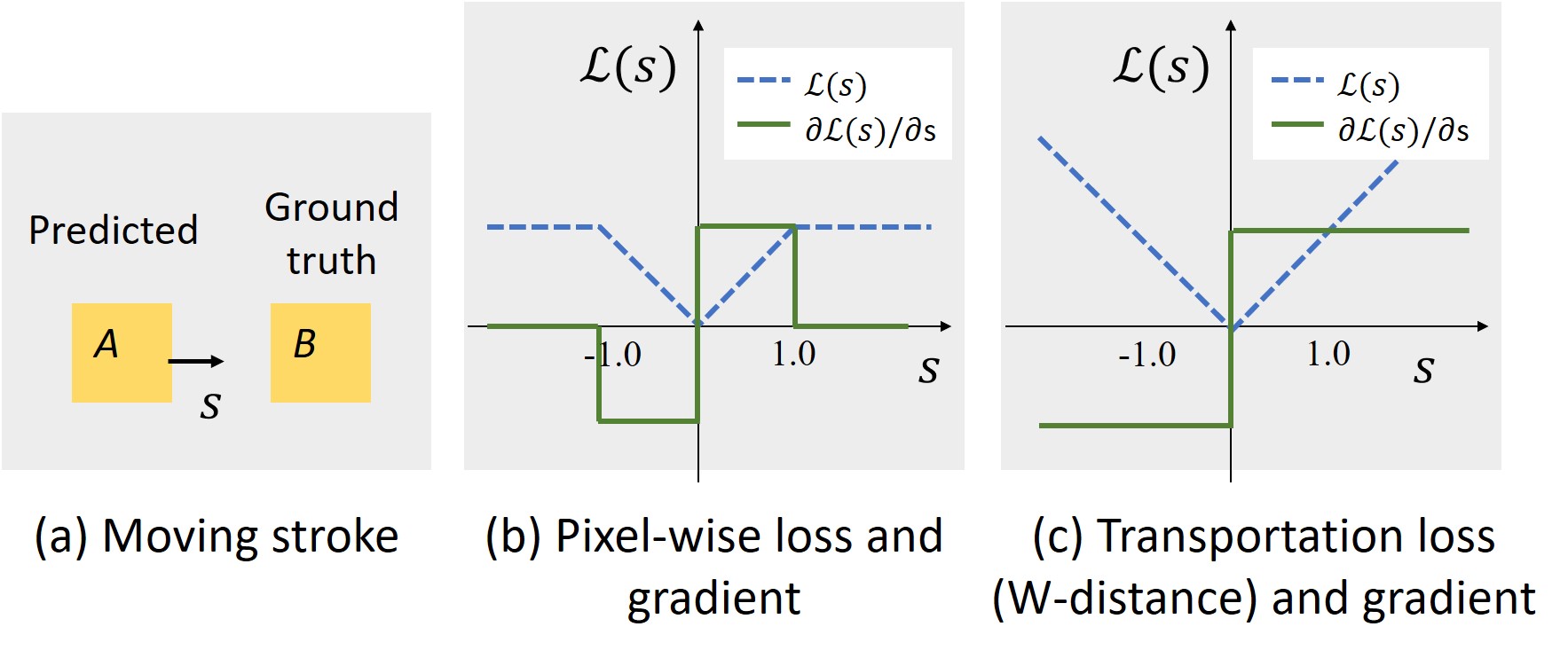}} \\
    \caption{A simple example to explain why pixel-wise loss $\mathcal{L}$ may have the zero-gradient problem ($\partial \mathcal{L} / \partial s=0$) during the stroke update. (a) When we move a square-shaped stroke $A$ along the $s$ direction to its target $B$, the pixel loss would remain constant when there is no overlap between $A$ and $B$. This will cause zero-gradients in the loss function, as shown in (b). As a comparison, the optimal transportation (OT) loss  in (c) does not have such a problem since the loss is correlated to the physical distance between $A$ and $B$.}
    \label{fig:gradient}
\end{figure}

There are many ways to define the similarity between the rendered output $h_T$ and the reference $\hat{h}$, and perhaps the most straight-forward one is to define as pixel-wise loss, e.g., $\ell_1$ or $\ell_2$ losses. Note that when we manipulate the image by directly optimizing from their pixel space, using the pixel-wise loss can work pretty well. However, when it comes to optimizing stroke parameters, we show that pixel loss does not always guarantee effective gradient descent. Particularly, when the rendered stroke and its ground truth do not share overlapped regions, there will be a zero-gradient problem. In Fig.~\ref{fig:gradient}, we give a simple example of why such cases would happen.

\begin{figure}[ht]
    \centering{\includegraphics[width=\linewidth]{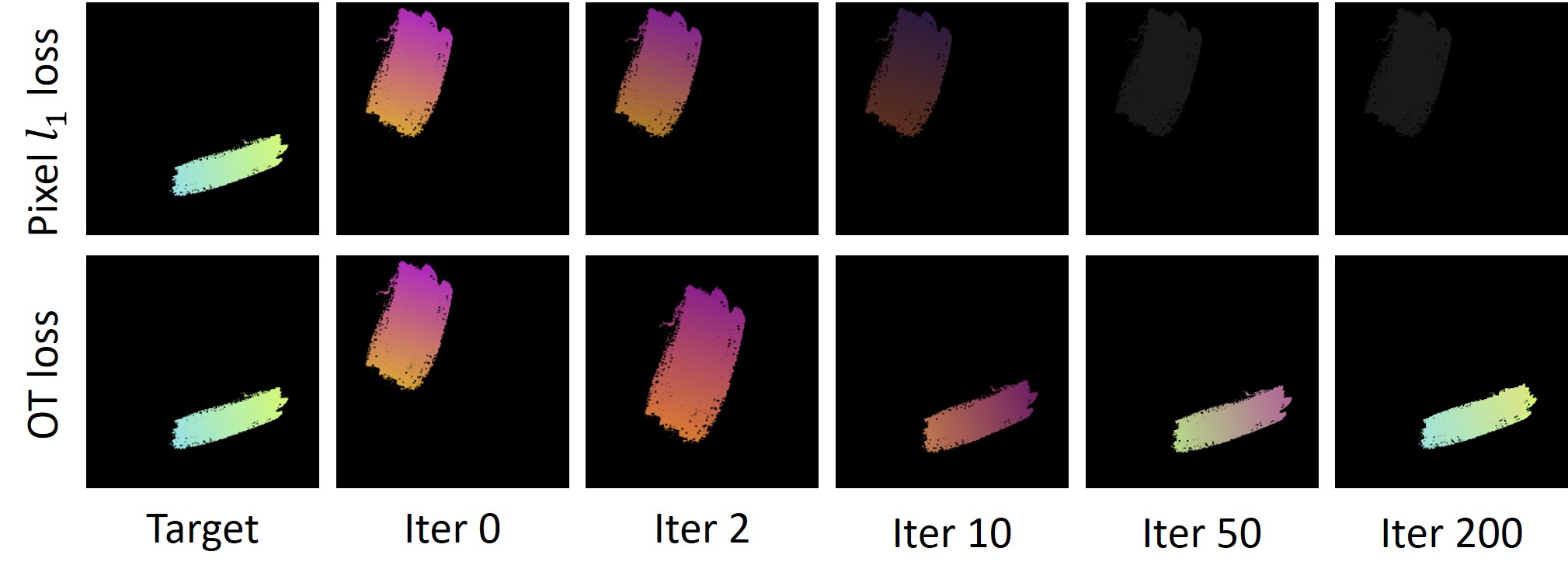}} \\
    \caption{A comparison between the pixel loss (1st row) and transportation loss (2nd row) on ``pushing'' a stroke from its initial state to its target. Using the proposed transportation loss, the stroke nicely converged to the target. As a comparison, the pixel loss fails to converge due to the zero-gradient problem in its position and scale.}
    \label{fig:strokemoves}
\end{figure}

Suppose we have a square-shaped stroke $A$ generated by a renderer and we aim at optimizing its parameters to make it converge to the target location $B$.  If we move $A$ along the horizontal direction and we suppose this movement is controlled by a certain parameter dimension (e.g., center x-coordinate). Clearly, when there is no overlap between the $A$ and $B$, the sum of the pixel loss between the canvas and the ground truth will remain a constant, which leads to a zero gradient and fails to guide the stroke update. Fig.~\ref{fig:gradient} (b) shows the pixel loss and its gradient on the moving distance. This problem can be effectively solved by re-defining the loss as the amount of transportation effort spent on the movement, where the farther $A$ and $B$ are, the greater the effort required, as shown in Fig.~\ref{fig:gradient} (c). In Fig.~\ref{fig:strokemoves}, we give another visualization example of how strokes behave when optimizing the pixel-loss and the transportation loss. We see that with pixel $\ell_1$ loss, the stroke fails to move along the right direction since there is no gradient on its parameterized locations while using transportation loss makes the stroke nicely converges to the target.

\begin{figure}[t]
    \centering{\includegraphics[width=\linewidth]{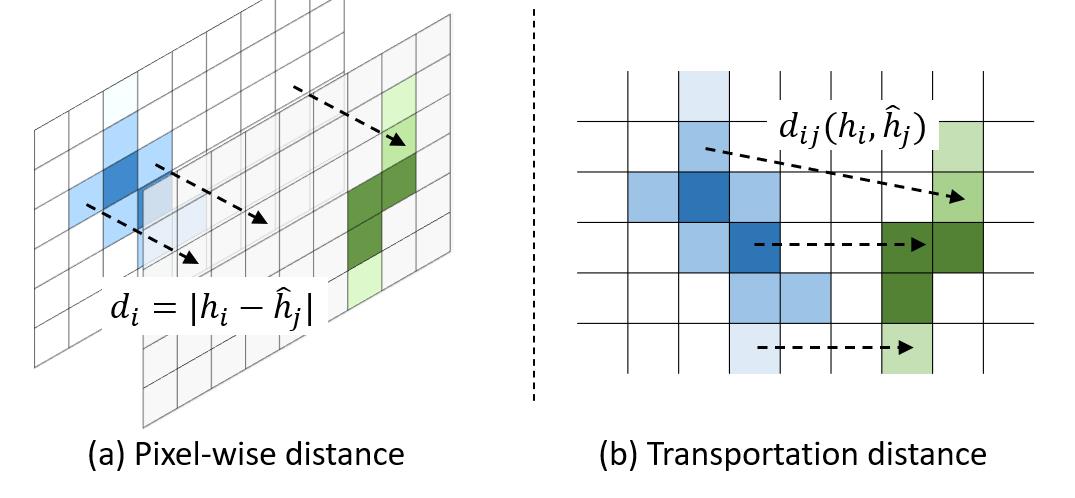}} \\
    \caption{A comparison between the pixel-wise distance and the transportation distance.}
    \label{fig:visualize_pxl_ot_distance}
\end{figure}

\subsection{Optimal transport for stroke searching}

We define the minimum transportation efforts, i.e., the Wasserstein distance, as an effective measure of similarity loss between the canvas and the reference image. Fig.~\ref{fig:l1_vs_ot_rendered} gives a brief visualization of how optimal transport loss differs from pixel loss on measuring the image similarity.

Given a rendered canvas $h$ and a reference image $\hat{h}$, we define their normalized pixel values $\bm{p}_h$ and $\hat{\bm{p}}_h$ as their probabilistic marginal functions. Here we omit the subscript $T$ for simplicity. We define $\mathbf{P}\in\mathbb{R}_+^{n\times n}$ as the joint probability matrix whose $(i,j)$-th element denotes the joint probability of the $i$-th pixel in $h$ and $j$-th pixel in $\hat{h}$, where $n$ is the number of pixels in the image. We let $\mathbf D$ be the cost matrix whose $(i,j)$-th element denotes the Euclidean distance between the $i$-th pixel's location in $h$ and $j$-th pixel's location in $\hat{h}$. Thus, the matrix $\mathbf{D}$ list all the labor costs of moving a ``unit pixel mass'' from one position in $h$ to another one in $\hat{h}$. In the discrete case, the classic optimal transport distance can be written as a linear optimization problem $\min_{\mathbf{P}\in \mathcal{U}}\langle \mathbf{D}, \mathbf{P} \rangle$, where $\mathcal{U} :=\{\mathbf{P} \in \mathbb{R}_+^{n\times n} \ |  \ \mathbf{P}\bm{1}_n = \bm{p}_h,\mathbf{P}^T\bm{1}_n = \hat{\bm{p}}_h\}$.

In this paper, we employ a smoothed version of the classic optimal transport distance with an entropic regularization term, which yields the celebrated Sinkhorn distance~\cite{cuturi2013sinkhorn}. The Sinkhorn distance is differentiable \cite{luise2018differential} and
enjoys benign mathematical properties that can result in much lower computational cost than the original one \cite{cuturi2013sinkhorn}. The primary idea is to consider an extra entropic constraints on the joint probability matrix $\mathbf{P}$ apart from $\mathcal{U}$. Further using the Lagrange multiplier, one can transform the problem into a regularized form as \eqref{eq:reg}. We let $\mathcal{L}_{ot}$ be the optimal transportation loss, namely the minimum transportation effort on moving strokes from one location to another, and then define it as the Sinkhorn distance in the following way
\begin{align}
\begin{aligned}\label{eq:reg}
    &\mathcal{L}_{ot}(h,\hat{h}) := \langle \mathbf{D}, \widetilde{\mathbf{P}}_{\lambda} \rangle, \text{ with }\\
    &\widetilde{\mathbf{P}}_{\lambda} = \mathop{\mathrm{argmin}}_{\mathbf{P}\in \mathcal{U}}~\langle \mathbf{D}, \mathbf{P} \rangle-\frac{1}{\lambda} E(\mathbf{P}),
\end{aligned}
\end{align}
where the entropy $E(\mathbf{P}) :=- \sum_{i,j=1}^n \mathbf{P}_{i,j} \log \mathbf{P}_{i,j}$.

The optimized transport loss can thus be easily integrated to the parameter searching pipeline and can be easily optimized together with other losses. We finally define the similarity loss as a combination of pixel $\ell_1$ loss and the optimal transportation loss:
\begin{equation}\label{eq:loss}
    \mathcal{L} = \beta_{\ell_1}\mathcal{L}_{\ell_1} + \beta_{ot}\mathcal{L}_{ot},
\end{equation}
where $\beta$'s control the balance of the above two objectives.

\begin{figure*}
    \centering{\includegraphics[width=1.0\linewidth]{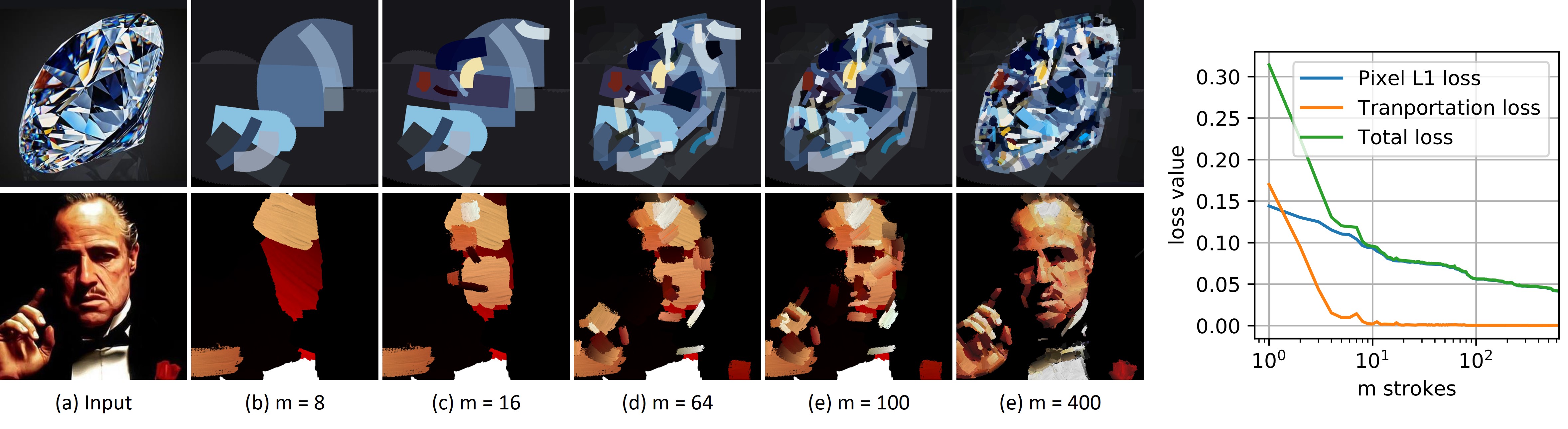}} \\
    \caption{Stroke-by-stoke painting result of our method with marker-pen (1st row) and oil-paint brush (2nd row). On the right, we also plot the loss curves (pixel $\ell_1$ loss, transportation loss, and total loss) as the painting proceeded.}
    \label{fig:stroke_by_stroke}
\end{figure*}

\begin{figure}
    \centering{\includegraphics[width=1.0\linewidth]{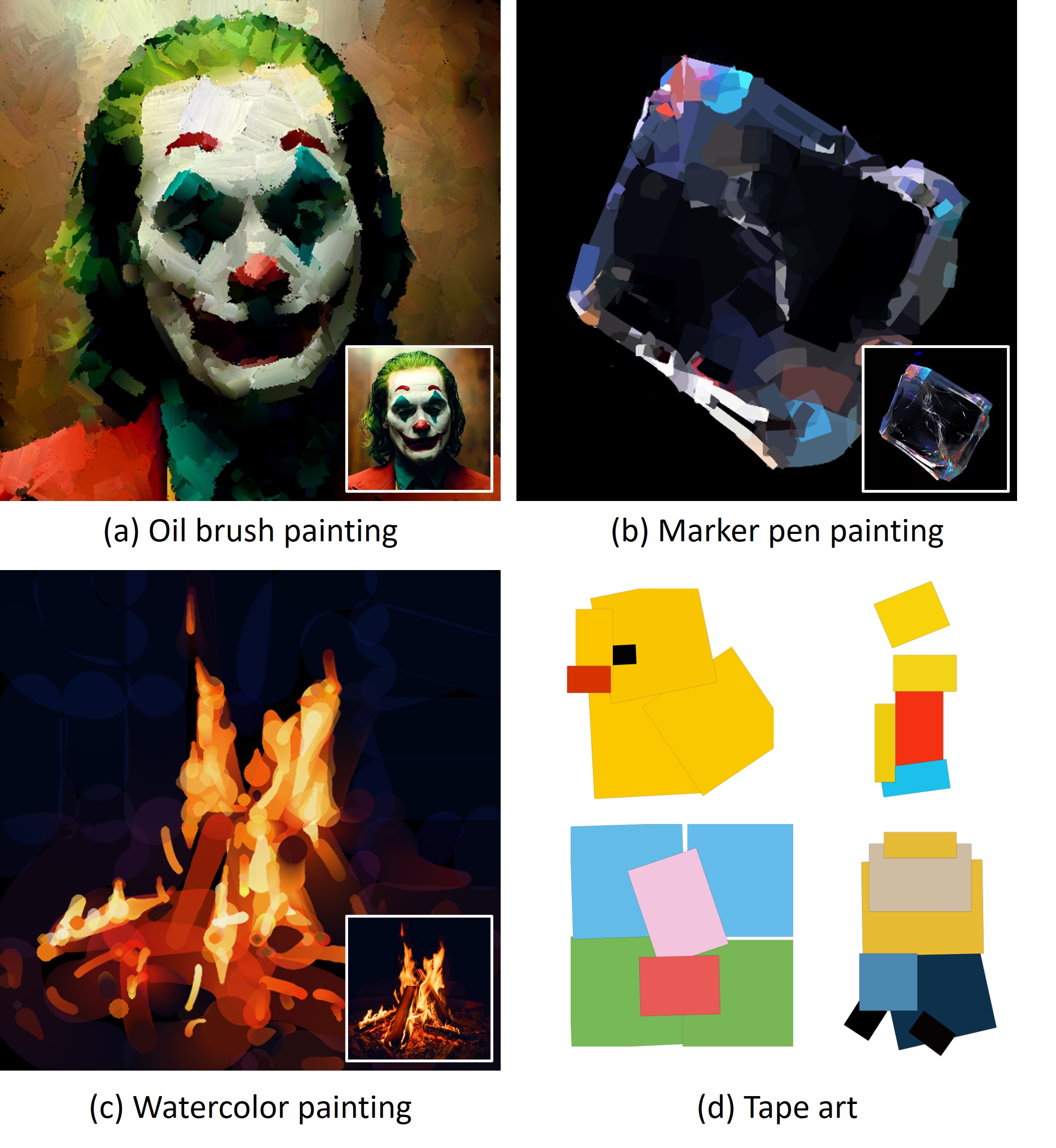}} \\
    \caption{(a)-(c) Stylized paintings generated by our method. In (d), we also show some highly abstract tape arts of cartoon characters generated by our method. Can you guess who they are? (See answers below)}
    \label{fig:rst}
\end{figure}

\begin{figure*}
    \centering{\includegraphics[width=0.95\linewidth]{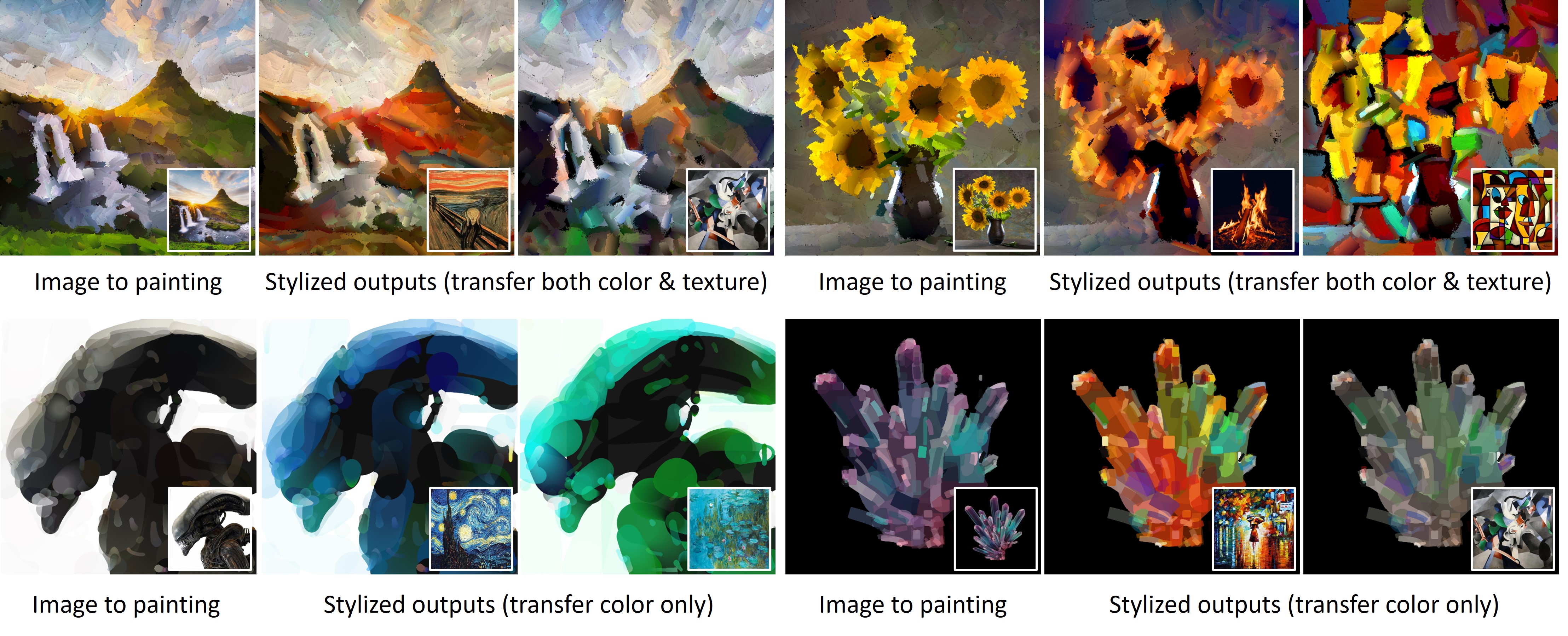}}\\
    \caption{Style transfer results by using our method. The 1st row shows the results that transfer the style of both color and texture. The 2nd row shows the results that transfer color only.}
    \label{fig:nst}
\end{figure*}

\subsection{Joint optimization with neural style transfer}

Since we frame our stroke prediction under a parameter searching paradigm, our method naturally fits the neural style transfer framework. Neural style transfer models are typically designed to updating image pixels to minimize the combination of a content loss and a style loss~\cite{gatys2016image}
$\mathcal{L}_{nst} = \mathcal{L}_{ctt} + \mathcal{L}_{sty}$,
where $\mathcal{L}_{ctt}$ and $\mathcal{L}_{sty}$ correspond to the constraints on the image content and style respectively.

To edit the global visual style of the rendering output, we extend the similarity loss (\ref{eq:loss}) with the style loss above, and define the extended similarity loss as follows:
\begin{equation}\label{eq:loss_extended}
    \mathcal{L} = \beta_{\ell_1}\mathcal{L}_{\ell_1} + \beta_{ot}\mathcal{L}_{ot}  + \beta_{sty}\mathcal{L}_{sty},
\end{equation}
where $\beta$'s are the weights to balance the losses. We follow Gatys \etal~\cite{gatys2016image} and compute the style loss as the square error of the Gram matrices of the features extracted by deep CNNs. We use VGG-19~\cite{simonyan2014very} as the feature extraction network and compute the Gram matrices on the features produced by the 2nd, 4th, and 7th convolutional layers.

\subsection{Implementation Details}

{\bf Network architecture.} We build our shading network similar to DCGAN's generator~\cite{radford2015unsupervised}, which consists of six transposed conv layers. We remove the Tanh activation from the output layer and observe a better convergence. In our rasterization network, we follow Huang \etal~\cite{huang2019learning} and first build a positional encoder with four fully-connected layers then build a pixel decoder with six conv layers and three pixel-shuffle layers~\cite{shi2016real}. We also experiment with the architecture of ``UNet'' \cite{ronneberger2015u}, wherein this case, we tile the stroke parameters on their spatial dimensions to a 3D tensor as the input of the network.

{\bf Training details.} We train our renderer by using Adam optimizer~\cite{kingma2014adam}. We set batch size to 64, learning rate to 2e-4, and betas to (0.9, 0.999). We reduce the learning rate to its 1/10 every 100 epochs and stop training after 400 epochs. In each epoch, we randomly generate 50,000x64 ground truth strokes using a vector engine. We set the rendering output size to 128x128 pixels. We train renderers separately for each stroke type.

{\bf Progressive rendering.} To render with more details, we design a progressive rendering pipeline in both the scale and action dimension. We first start from searching parameters on a single 128x128 canvas and then divide the canvas into $m\times m$ blocks ($m=2, 3, 4, ...$) with overlaps and search on each of them accordingly. In each block scale, we gradually add new strokes to an ``active set'' and update the strokes progressively. In each update, we optimize all strokes within the active set at the same time. We run gradient descent for $20\times N$ steps for each block, where $N$ is the number of strokes in each block, same for all blocks regardless of their scales.

\footnotetext{Answer: Rubber Duck, Bart Simpson, Peppa Pig, and, Minions.}

{\bf Other details.} We set $\beta_{\ell_1}=1.0$ and $\beta_{ot}=0.1$. We use the RMSprop~\cite{hinton2012neural} optimizer for gradient descent with learning $\mu=0.01$. In the style transfer, we set $\beta_{sty}=0.5$ and update for 200 steps. In Sinkhorn loss, we resize the canvas to 48x48 pixels to speed up the computation. We set the entropic regularization term $\epsilon=0.01$, and the number of steps in sinkhorn loop $n_{iter}=5$. For more details on the network architectures and stroke parameterization, please refer to our supplementary material.

\begin{figure}
    \centering{\includegraphics[width=0.95\linewidth]{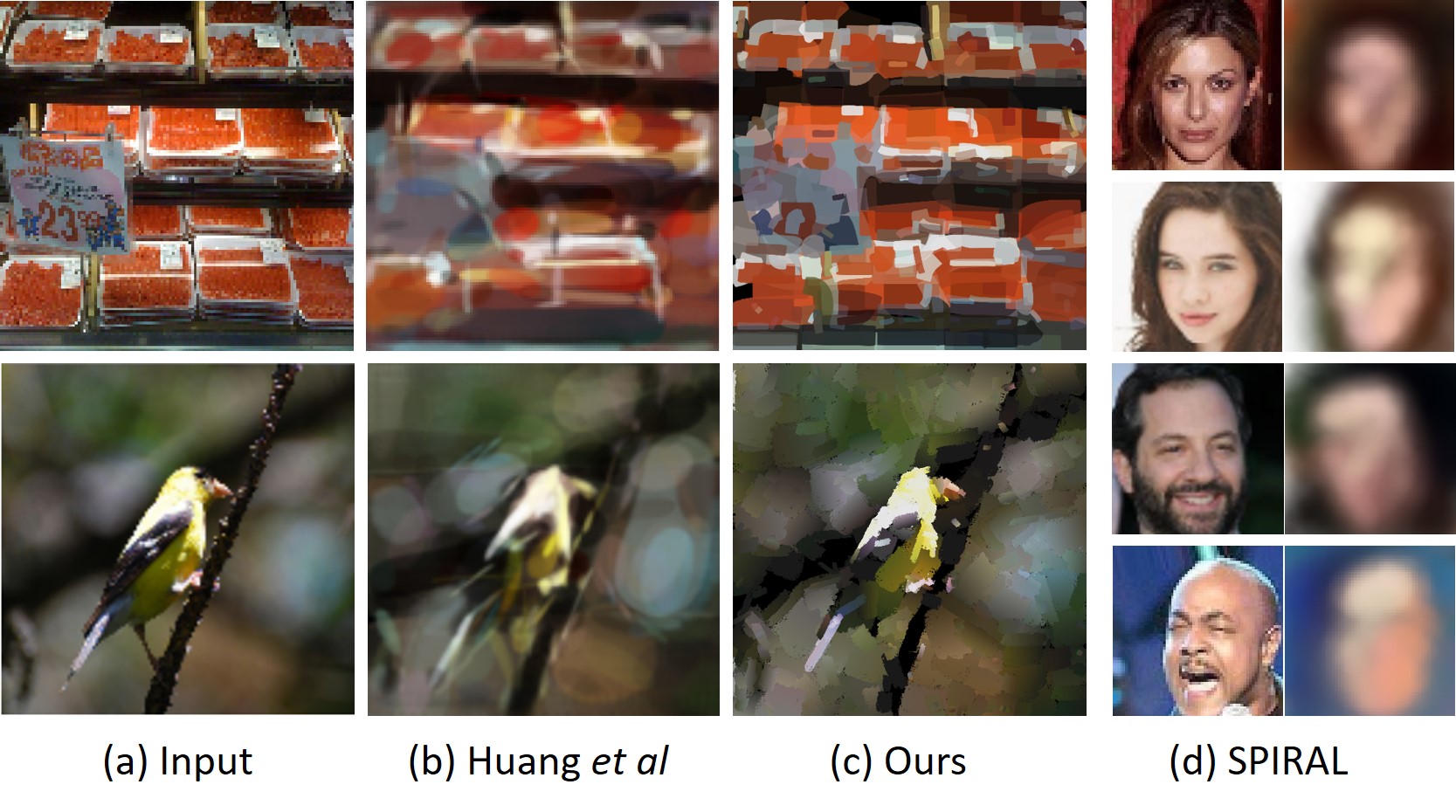}} \\
    \caption{A comparison of the paintings created by our method (400 strokes), ``Learning-to-Paint''~\cite{huang2019learning} (400 strokes), and SPIRAL~\cite{ganin2018synthesizing} (20 strokes). The results in (b) and (d) are from their papers.}
    \label{fig:haung_cmp}
\end{figure}

\begin{figure}
    \centering{\includegraphics[width=0.90\linewidth]{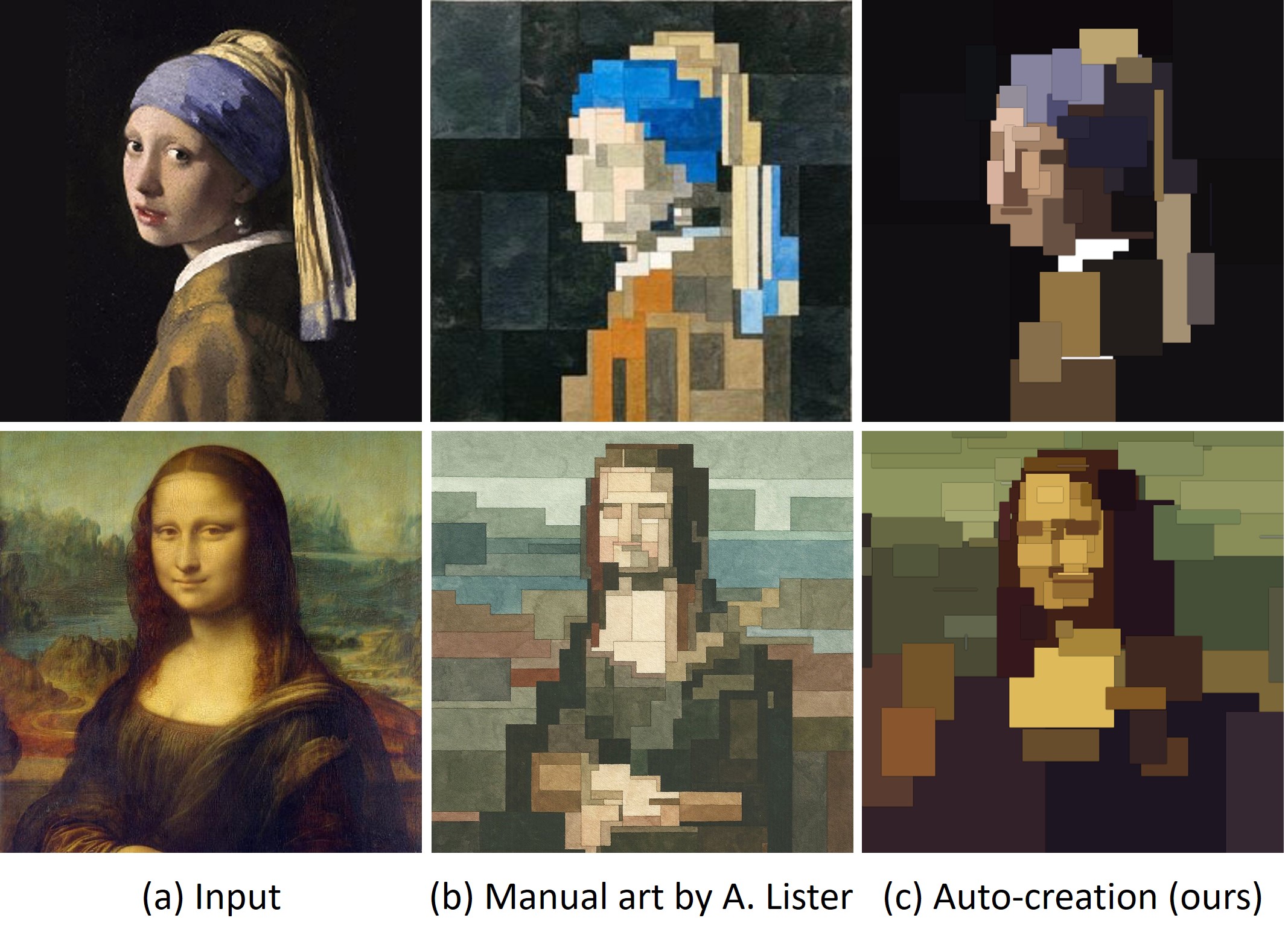}} \\
    \caption{Artworks created manually by a New York artist \href{https://adam-lister-gallery.myshopify.com/}{Adam Lister} and created automatically by our method.}
    \label{fig:adam_lister}
\end{figure}


\section{Experimental Analysis}

\subsection{Stylized painting generation}

Fig.~\ref{fig:rst} shows a group of stylized paintings generated by using our method with different stroke types. In (d), we also show several tape-artworks of well-known characters that are automatically created by our method with a minimum number of strokes. We can see that our method successfully learns high-level abstractions of the characters and vividly portrays their shape and color. Fig.~\ref{fig:stroke_by_stroke} shows the stroke-by-stroke painting results by using different stroke brushes. On the right of this figure, we also plot the changes in the loss values as the painting proceeded. We can see that in the very first few drawing steps, our method nicely captures the global appearance of the object, and the drawing then gradually goes from macro to detail. Fig.~\ref{fig:nst} shows more examples of our painting results as well as their style transfer results. We can see by integrating the style loss in (\ref{eq:loss_extended}), both color and textures can be successfully transferred to the paintings with their content remaining unchanged.

\begin{figure}
    \centering{\includegraphics[width=0.9\linewidth]{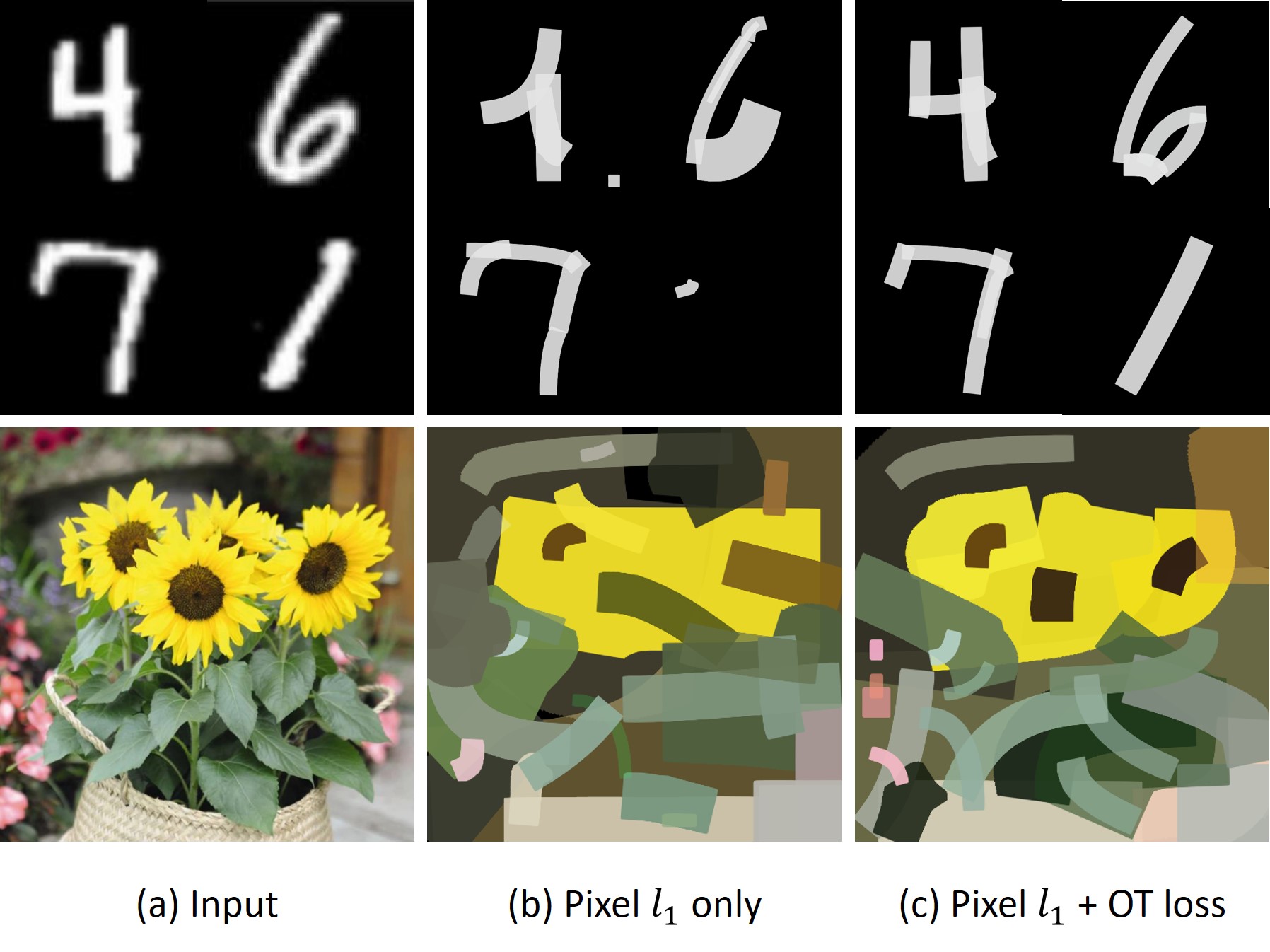}} \\
    \caption{A comparison between the strokes optimized by using pixel $\ell_1$ loss only, and pixel $\ell_1$ loss + optimal transportation loss on a flower picture and MNIST digits~\cite{lecun1998gradient}. The transportation loss can help recover more details of the image, especially when the stroke is initialized with no overlap with the target region.}
    \label{fig:l1_vs_ot_rendered}
\end{figure}

\subsection{Comparison with other methods}

In Fig.~\ref{fig:haung_cmp}, we compare our method with two recent proposed stroke-based image-to-painting translation methods: 1) ``Learning-to-Paint''~\cite{huang2019learning}, and 3) ``SPIRAL''~\cite{ganin2018synthesizing}, where both of them trains RL agent to paint. We can see our method generates more vivid results with a clear distinction on brush textures, while other methods tend to produce blurred results. We also compare the stylized artworks created by our method with those created manually. Fig.~\ref{fig:adam_lister} shows the comparison, where the second column shows the results create by a famous artist Adam Lister from New York. Those manual artworks are from his \href{https://adam-lister-gallery.myshopify.com/}{official gallery website.}
We can see both the manual result and automated results present a low-bit artistic geometry painting effect.

\subsection{Controlled Experiments}

\textbf{Pixel loss vs transportation loss.} To evaluate the effectiveness of our transportation loss function, we design the following experiment where we visually compare the painting generated by 1) pixel $\ell_1$ loss only, and 2) pixel $\ell_1$ loss + transportation loss. Fig.~\ref{fig:l1_vs_ot_rendered} shows the comparison result. We test on a flower picture and several digits from the MNIST dataset~\cite{lecun1998gradient}. We can see that transportation loss helps to recover more details. Particularly, in the digit ``1'' and digit ``4'', using only the pixel $\ell_1$ loss may produce bad convergence results on stroke parameters, which is similar to the failure case we have shown in Fig.~\ref{fig:strokemoves}.

\begin{figure}
    \centering{\includegraphics[width=\linewidth]{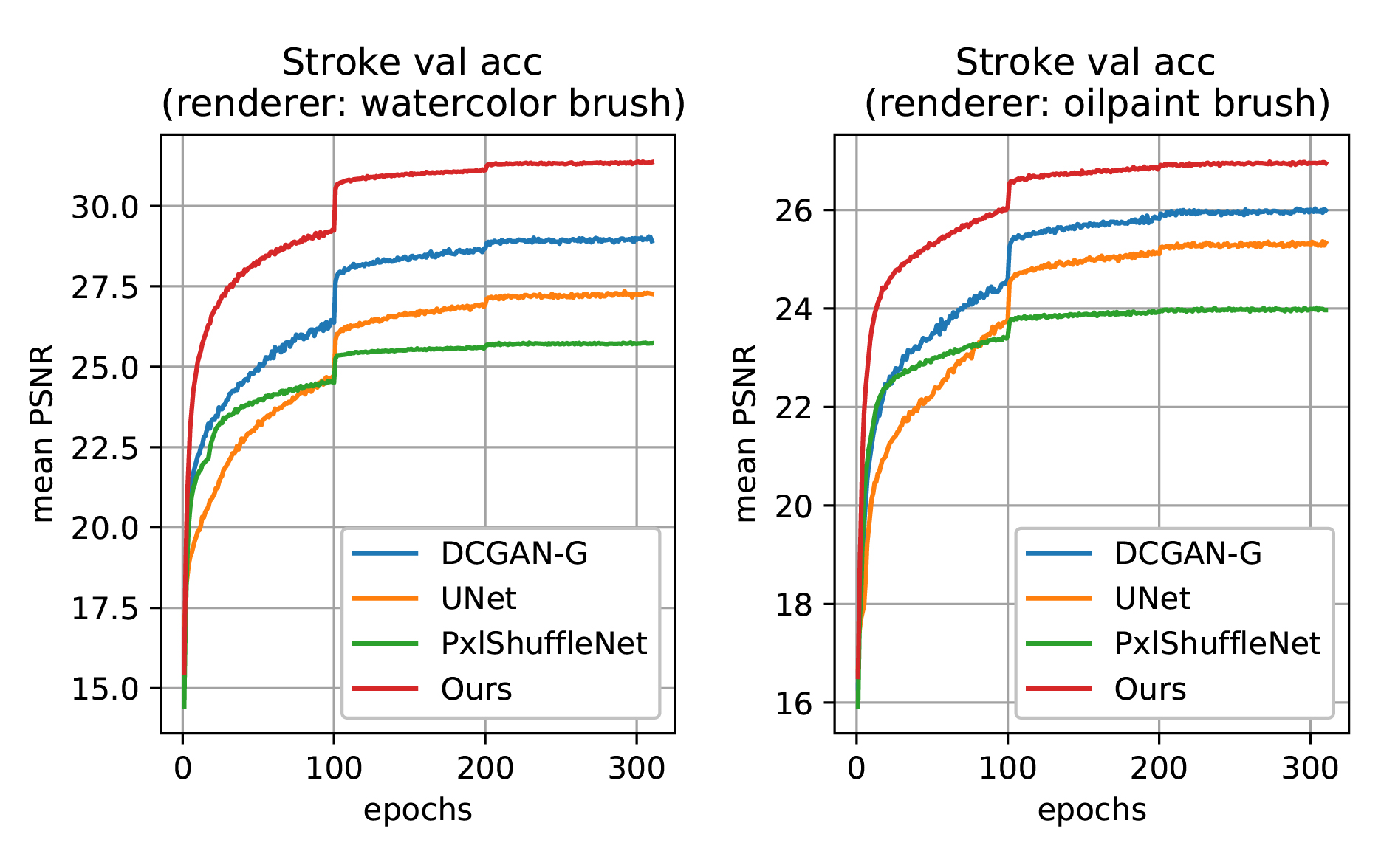}} \\
    \caption{Validation accuracy (mean PSNR of the rendered foreground and alpha matte) of different neural renderers: DCGAN-G~\cite{shi2019face,zhu2016generative}, UNet~\cite{ronneberger2015u}, PxlShuffleNet~\cite{huang2019learning}. Our render outperforms other renders with a large margin in both accuracy and convergence speed.}
    \label{fig:network_acc_curves}
\end{figure}

\begin{figure}
    \centering{\includegraphics[width=\linewidth]{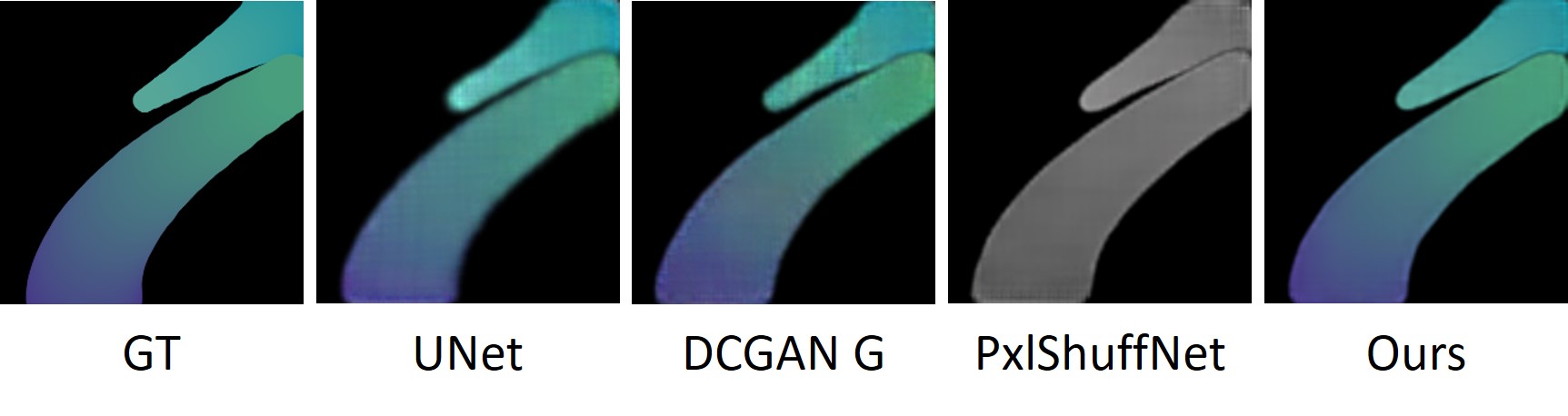}} \\
    \caption{A visual comparison between the results rendered by different neural renderers: UNet~\cite{ronneberger2015u}, DCGAN-G~\cite{shi2019face,chen2020neural},  PxlShuffleNet~\cite{huang2019learning}, and ours.}
    \label{fig:network_stroke_cmp}
\end{figure}

\textbf{Compare with other neural renderers.} We compare our proposed renderer with some recent ones, including 1) a DCGAN-Generator-like renderer, which has been recently used for rendering 3D objects and faces~\cite{shi2019face,zhu2016generative,shi2020neural,chen2020neural}, and 2) a positional encoder + decoder architecture which has been used for rendering sketches and paintings~\cite{huang2019learning,zheng2018strokenet}. Here we refer to the former one as ``DCGAN-G'' and refer to the latter one as ``PixShuffleNet''. We compare their accuracy on randomly generated strokes. Fig.~\ref{fig:network_acc_curves} shows the changes in their validation accuracy on different training epochs. We test on the watercolor brush and the oil-paint brush rendering tasks separately, but the marker-pen and tape-art renderer also have similar performances. All the renderers are trained under the same settings. We can see that in all brush settings, our neural renderer achieves the highest accuracy and is much better than other renderers (+2.5$\sim$+5.0 higher than others in PSNR). In addition, our renderer also has a much faster convergence speed than. In Fig.~\ref{fig:network_stroke_cmp}, we made a visual comparison between the different renderers. Our method produces the best result in terms of both color faithness and the high-frequency details. We also notice that the PxlSuffleNet fails to recover the color although we do have trained it on the color space. This could be caused by the highly coupled color and shape representations of its positional encoding or pixel-shuffle layers.

\textbf{Rasterization network and shading network.} In this experiment, we evaluate the effectiveness of the two components in our neural renderer: the rasterization network and the shading network. We design the following experiment where we first test on each of them separately and then test on our full implementation. All the networks are trained under the same settings. Table~\ref{tab:ablation} shows the comparison results. We observe a noticeable accuracy drop (mean PSNR of the foreground and alpha matte) when we remove either of the two networks from our renderer, which suggests the effectiveness of the dual-pathway design of our renderer.

\begin{table}\small
\begin{center}
\begin{tabular}{l|cc}
\toprule
Renderer/Stroke  & Oil-paint & Watercolor \\
\midrule
Ours (rasterization only) & 24.015 & 25.769 \\
Ours (shading only) &  26.048 & 29.045 \\
Ours (rasterization + shading) & 26.982 & 31.389 \\
\bottomrule
\end{tabular}%
\end{center}
\vspace{-1.5em}
\caption{Ablation study of the proposed neural renderer.}
\label{tab:ablation}%
\end{table}%

\section{Conclusion}

We explore the nature of human painting by using differentiable stroke rendering. We consider this artistic creation process under a stroke parameter searching paradigm that maximizes the similarity between the sequentially rendered canvas and the reference image. Our method can generate highly realistic and painting artworks in vector format with controllable styles. We deal with the image similarity measurement from the perspective of optimal transportation and tackle the disentanglement of the color and shape with a dual-pathway neural renderer. Controlled experiments suggest the effectiveness of our design.

{\small
\bibliographystyle{ieee_fullname}
\bibliography{egbib}
}

\clearpage

\section{Appendix}
\subsection{Detailed configurations of our neural renderer}

In Table~\ref{tab:shading_net} and In Table~\ref{tab:rasterization_net}, we show a detailed configuration of our rasterization network and our shading network. Specifically, in a $c\times w \times w / s$ layer, $c$ denotes the number of filters, $w \times w$ denotes the filter's size and $s$ denotes the stride size. The output size is formatted as height $\times$ width $\times$ channel.

\begin{table}[h]\small
\centering
\caption{Details of our shading network.}
\begin{tabular}{l|ccc}
    \toprule
      & \textbf{Layer} & \textbf{Config} & \textbf{Out size}\\
    \midrule
     C1 & Deconv + BN + ReLU & 512$\times$4$\times$4~/~1 & 4$\times$4$\times$512\\
     C2 & Deconv + BN + ReLU & 512$\times$4$\times$4~/~2 & 8$\times$8$\times$512\\
     C3 & Deconv + BN + ReLU & 256$\times$4$\times$4~/~2 & 16$\times$16$\times$256\\
     C4 & Deconv + BN + ReLU & 128$\times$4$\times$4~/~2 & 32$\times$32$\times$128\\
     C5 & Deconv + BN + ReLU & 64$\times$4$\times$4~/~2 & 64$\times$64$\times$64\\
     C6 & Deconv + BN + ReLU & 3$\times$4$\times$4~/~2 & 128$\times$128$\times$3\\
    \bottomrule
\end{tabular}\label{tab:shading_net}
\end{table}

\begin{table}[h]\small
\centering
\caption{Details of our rasterization network.}
\begin{tabular}{l|ccc}
    \toprule
      & \textbf{Layer} & \textbf{Config} & \textbf{Out size}\\
    \midrule
     F1 & Full-connected + ReLU & 512 & 512\\
     F2 & Full-connected + ReLU & 1024 & 1024\\
     F3 & Full-connected + ReLU & 2048 & 2048\\
     F4 & Full-connected + ReLU & 4096 & 4096\\
     V1 & View & - & 16$\times$16$\times$16\\
     C1 & Conv + ReLU & 32$\times$3$\times$3~/~1 & 16$\times$16$\times$32\\
     C2 & Conv + Shuffle & 32$\times$3$\times$3~/~2 & 32$\times$32$\times$8\\
     C3 & Conv + ReLU & 16$\times$3$\times$3~/~1 & 32$\times$32$\times$16\\
     C4 & Conv + Shuffle & 16$\times$3$\times$3~/~2 & 64$\times$64$\times$4\\
     C5 & Conv + ReLU & 8$\times$3$\times$3~/~1 & 64$\times$64$\times$8\\
     C6 & Conv + Shuffle & 4$\times$3$\times$3~/~2 & 128$\times$128$\times$1\\
    \bottomrule
\end{tabular}\label{tab:rasterization_net}
\end{table}

\subsection{Stroke parameterization}

We design four types of painting brushes: oil-painting brush, watercolor ink, marker pen, and color tapes. In Table~\ref{table:oilpaint-stroke-config},~\ref{table:markerpen-stroke-config},~\ref{table:watercolor-stroke-config}, and~\ref{table:colortape-stroke-config}, we give a  detailed description of each of the stroke parameter, where ``Controllers'' shows how the parameters of each stroke are configured, and $N$ represents the total number of parameters in each stroke.

For the ``marker pen'' and the ``watercolor ink'', we design the main trajectory of the stroke movement as a quadratic Bezier curve.
The shape of the Bezier curve is specified by three control points $P_0=(x_0, y_0)$, $P_1=(x_1, y_1)$, and $P_2=(x_2, y_2)$. Formally, the stroke is defined as
\begin{equation}
    B(t)=(1-t)^2P_0+2(1-t)tP_1+t^2P_2,
\end{equation}
where $0 \leq t \leq 1$. We define another set of parameters to control the stroke thickness and color. For the watercolor ink, we define thickness ($r_0, r_2$) and colors ($R_0, G_0. B_0, R_2, G_2, B_2$) separately at $P_0$ and $P_2$, while for marker pen, we use constant thickness $d$ and color ($R, G, B$) in each stroke trajectory.

For the ``color-tapes'', we define it as a solid color rectangle with a rotation angle $\theta\in [0, 180\degree]$. The position and size are define by ($x_0, y_0, h, w$).

For the ``oil-painting brush'', we define its parameters similar to the color-tapes. The shape parameters include position, size, and orientation. The colors are defined at the head and tail of the stroke separately ($R_0, G_0. B_0, R_2, G_2, B_2$). Since oil paints are not transparent, we ignore the transparency parameter $A$ and set it to a constant $A=1$. We also blend a texture map on top of the rendered stroke but we simply treat the texture as a constant map which is not updated during the parameter searching.

\begin{table}\small
    \centering
    \caption{Detailed parameterization of oil-painting strokes.}
    \begin{tabular}{l|l|c}
        \toprule
          & Controllers &  N \\
        \midrule
        shape & $x_0, y_0, h, w, \theta$ & 5\\
        color & $R_0, G_0, B_0, R_1, G_2, B_2$, ($A=1$) &  6\\
        \bottomrule
    \end{tabular}
    \label{table:oilpaint-stroke-config}
\end{table}

\begin{table}\small
    \centering
        \caption{Detailed parameterization of markerpen strokes.}
    \begin{tabular}{l|l|c}
        \toprule
          & Controllers &  N \\
        \midrule
        shape & $x_0, y_0, x_1, y_1, x_2, y_2, d$ & 7\\
        color & $R, G, B, A$  &  3\\
        \bottomrule
    \end{tabular}
    \label{table:markerpen-stroke-config}
\end{table}

\begin{table}\small
    \centering
        \caption{Detailed parameterization of watercolor ink.}
    \begin{tabular}{l|l|c}
        \toprule
          & Controllers &  N \\
        \midrule
        shape & $x_0, y_0, x_1, y_1, x_2, y_2, r_0, r_2$ & 8\\
        color & $R_0, G_0, B_0, R_1, G_2, B_2, A$  &  7\\
        \bottomrule
    \end{tabular}
    \label{table:watercolor-stroke-config}
\end{table}

\begin{table}\small
    \centering
        \caption{Detailed parameterization of the color-tapes.}
    \begin{tabular}{l|l|c}
        \toprule
          & Controllers &  N \\
        \midrule
        shape & $x_0, y_0, h, w, \theta$ & 5\\
        color & $R, G, B$, ($A=1$) &  3\\
        \bottomrule
    \end{tabular}
    \label{table:colortape-stroke-config}
\end{table}

\subsection{High resolution results}

\begin{figure*}[h]
    \centering{\includegraphics[width=0.95\linewidth]{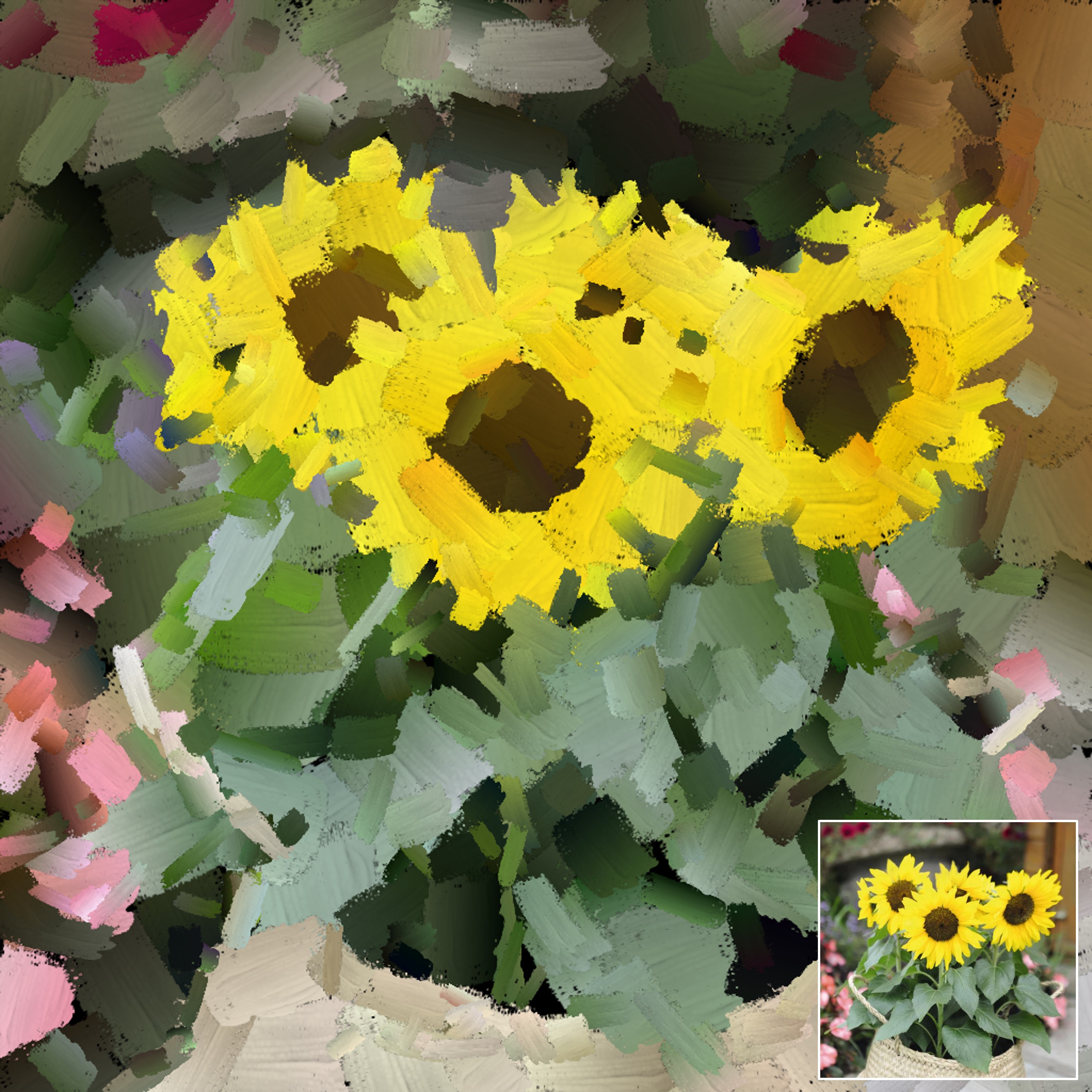}} \\
    \caption{A high resolution image-to-painting translation result of our method. The result is rendered at a resolution of 1024x1024 pixels.}
    \label{fig:supp_hd_rst1}
\end{figure*}

\begin{figure*}[h]
    \centering{\includegraphics[width=0.95\linewidth]{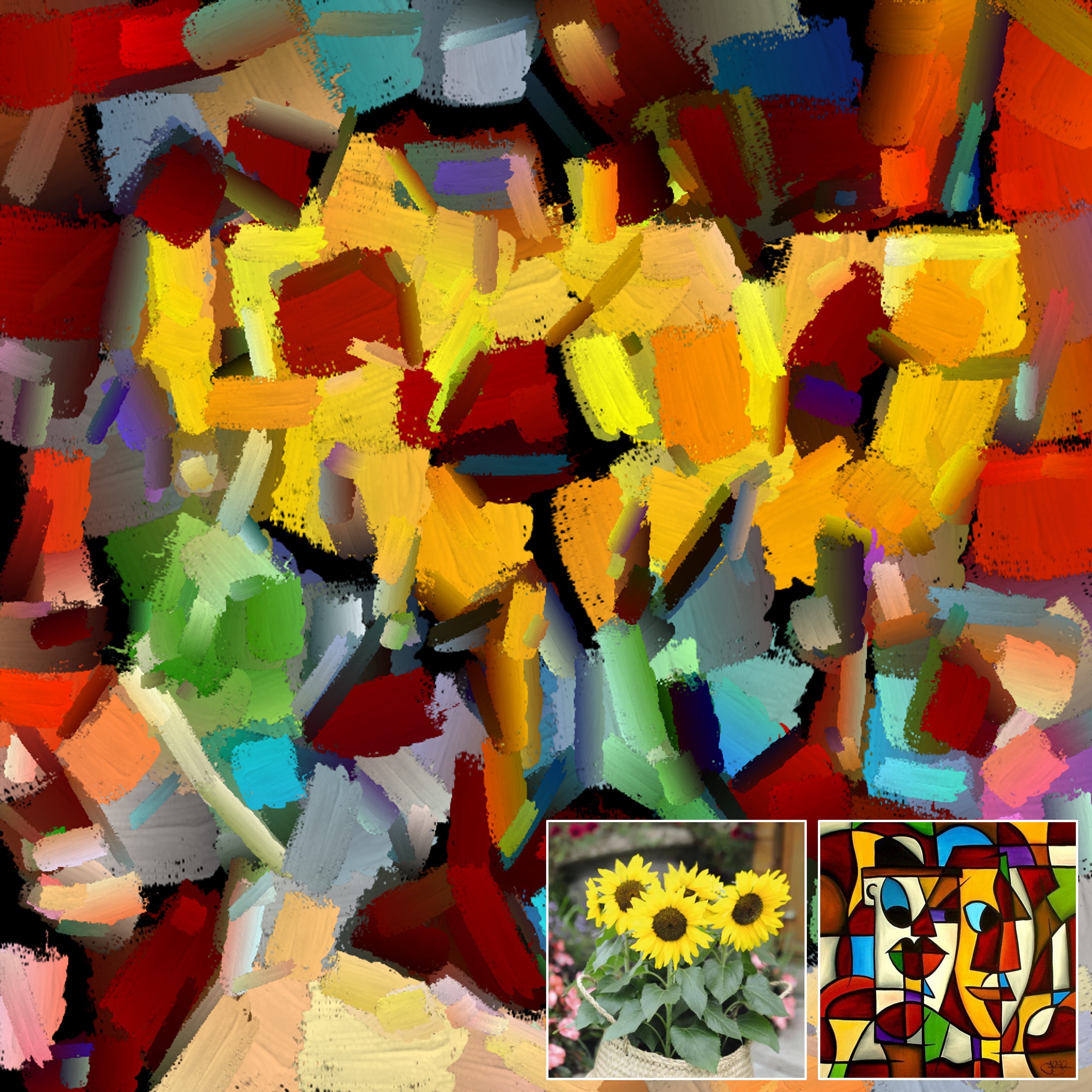}} \\
    \caption{A high resolution neural style transfer result of our method. The result is rendered at a resolution of 1024x1024 pixels.}
    \label{fig:supp_hd_rst2}
\end{figure*}

Since our painting results are generated with a vector format. We can render them at any resolutions. In Fig.~\ref{fig:supp_hd_rst1} and Fig.~\ref{fig:supp_hd_rst2}, we show two groups of results rendered at a 1024x1024 pixel resolution.

\end{document}